\documentclass[runningheads]{llncs}
\usepackage[T1]{fontenc}
\usepackage{amsmath}
\usepackage{amssymb}
\usepackage{tikz}
\usepackage{pgfplots}
\usepackage{float}
\usepackage{graphicx}
\usepackage{subcaption}
\usepackage{enumitem}
\usepackage{multirow}
\usepackage[misc]{ifsym}
\usepackage[marginal]{footmisc}

\usepackage{pifont}

\usepackage{graphicx}
\begin{document}
\title{SEDEG: Sequential Enhancement of Decoder and Encoder's Generality for Class Incremental Learning with Small Memory}

\titlerunning{SEDEG: Sequential Enhancement of Decoder and Encoder's Generality}

\author{Shaoling Pu\inst{1} \and
Hongyang Chen\inst{1}\thanks{Corresponding author.}  \and
Lingyu Zheng\inst{2} \and
Zhongwu Sun\inst{2}}
\authorrunning{S. Pu et al.}
%
\institute{Zhejiang Lab, Hangzhou,  China\\
\email{dr.h.chen@ieee.org,poorocky2@gmail.com}\and
Hangzhou Institute for Advanced Study, University of Chinese Academy of Sciences, China\\
\email{zhenglingyu22@mails.ucas.edu.cn, sunzhongwu23@mails.ucas.edu.cn}}
\maketitle              
\footnote{S. Pu and H. Chen -- Equal contributions.}
\footnote{H. Chen is with the Research Center for Scientific Data Hub, Zhejiang Lab, Hangzhou 311100, China (email: dr.h.chen@ieee.org; hongyang@zhejianglab.com).
This work is supported in part by National Key Research and Development Program of China 2022YFB4500300. The authors would also like to thank Dr.Kunzi Li for his valuable support and insightful suggestions.}

\begin{abstract}
In incremental learning, enhancing the generality of knowledge is crucial for adapting to dynamic data inputs. It can develop generalized representations or more balanced decision boundaries, preventing the degradation of long-term knowledge over time and thus mitigating catastrophic forgetting. Some emerging incremental learning methods adopt an encoder-decoder architecture and have achieved promising results.
 In the encoder-decoder achitecture, improving the generalization capabilities of both the encoder and decoder is critical, as it helps preserve previously learned knowledge while ensuring adaptability and robustness to new, diverse data inputs. However, many existing continual methods focus solely on enhancing one of the two components, which limits their effectiveness in mitigating catastrophic forgetting. And these methods perform even worse in small-memory scenarios, where only a limited number of historical samples can be stored. To mitigate this limitation, we introduces SEDEG, a two-stage training framework for vision transformers (ViT), focusing on sequentially improving the generality of both Decoder and Encoder. Initially, SEDEG trains an ensembled encoder through feature boosting to learn generalized representations, which subsequently enhance the decoder's generality and balance the classifier. The next stage involves using knowledge distillation (KD) strategies to compress the ensembled encoder and develop a new, more generalized encoder. This involves using a balanced KD approach and feature KD for effective knowledge transfer. Extensive experiments on three benchmark datasets show SEDEG's superior performance, and ablation studies confirm the efficacy of its components. The code is available at https://github.com/ShaolingPu/CIL.

\keywords{Incremental Learning  \and Decoder and Encoder’s Generality \and Vision Transformer \and Knowledge Distillation.}
\end{abstract}
\section{Introduction}

In the dynamic and open world, a learning system is required to possess the capability of incremental learning to adapt to new concepts. Unfortunately, deep neural networks (DNNs) often face the issue of catastrophic forgetting~\cite{mccloskey1989catastrophic}, which refers to the significant degradation of previously learned knowledge when updating the networks with new data. To address this issue, many approaches~\cite{rebuffi2016icarl,shin2017continual,hu2018overcoming,bang2021rainbow,li2022ckdf} have been proposed. These approaches can be categorized into two main types based on whether they store exemplars of previous tasks: (i) Exemplar-free approaches and (ii) Exemplar-based approaches. Exemplar-free approaches mainly consist of regularization-based methods~\cite{zeng2019continual,zhu2021class} and parameter isolation methods~\cite{singh2020calibrating,singh2021rectification}. However, regularization-based methods often suffer from over-regularization, leading
 to subpar performance.  In contrast,  exemplar-based approaches often integrate experience replay with various regularization techniques ~\cite{chaudhry2018efficient,chaudhry2020continual,wang2023improved} or knowledge transferring techniques~\cite{cha2021co2l,ahn2021ss,lin2022anchor,wang2022foster} to maintain previously learned knowledge.  Due to the storage of previous data, these methods often achieve superior performance. In this paper, we also concentrate on this kind of approach.
 
\begin{figure}[htbp]
    \centering
    \begin{minipage}{0.45\textwidth}
        \centering
        \includegraphics[width=\linewidth]{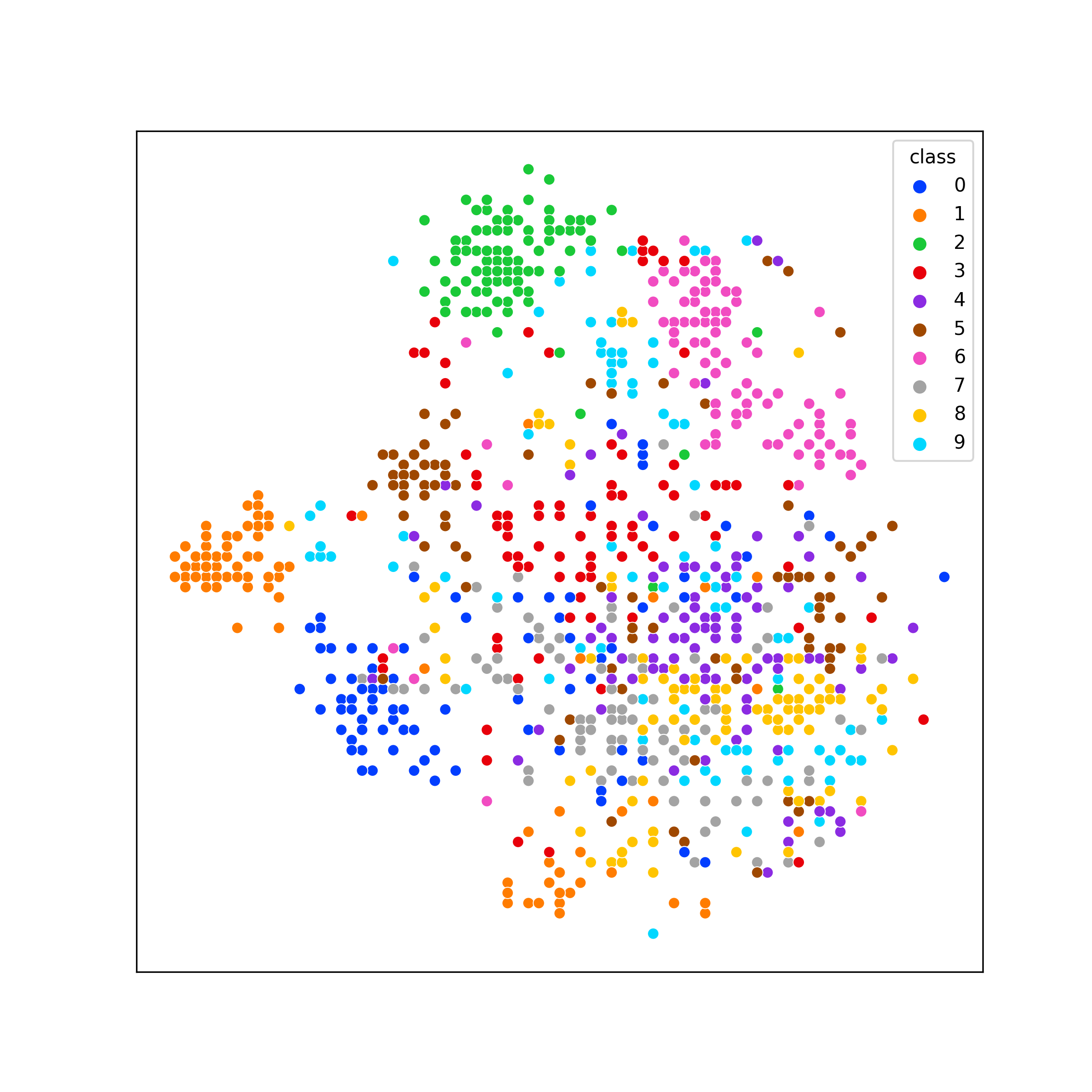}
        \subcaption{t-SNE map of DyTox.}
        \label{fig:dytox}
    \end{minipage}
    \hfill
    \begin{minipage}{0.45\textwidth}
        \centering
        \includegraphics[width=\linewidth]{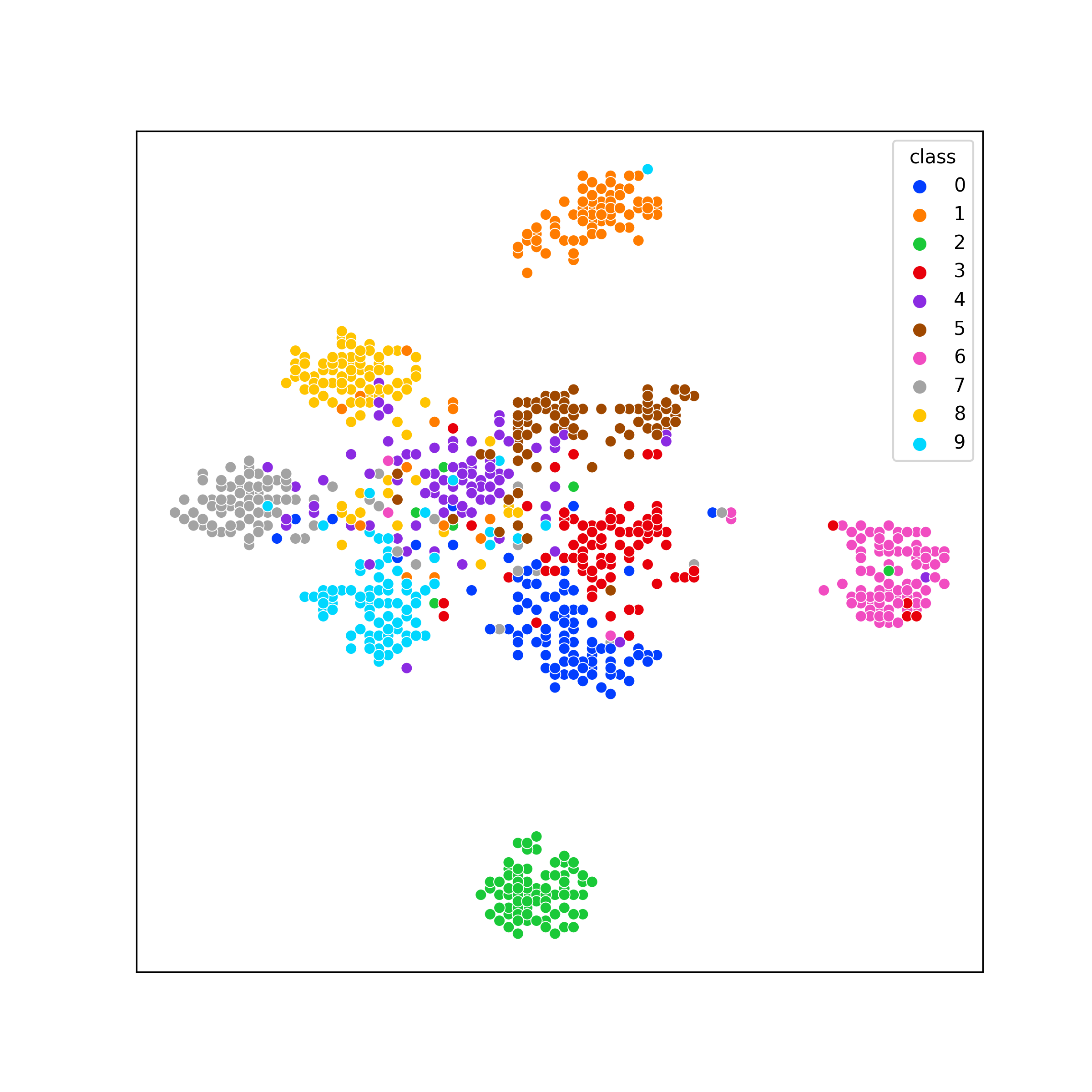}
        \subcaption{t-SNE map of SEDEG.}
        \label{fig:sedeg}
    \end{minipage}
    \caption{Comparison of t-SNE maps on evaluation dataset. (CIFAR-100~\cite{krizhevsky2009learning}, 20 tasks, \textbf{after the training of the second task}). (a) SEDEG effectively distinguishes clusters of different categories. (b) DyTox fails to effectively differentiate them.}
    \label{fig:tsne-maps}
\end{figure}

 In incremental learning~(IL), improving the generality of learned knowledge is paramount. This enhances adaptability and robustness to dynamic changes in data inputs. For example, Li et al.~\cite{li2022ckdf} and Wang et al.~\cite{wang2022foster} obtained more generalized representations using different strategies, both demonstrating that such representations prevent the degradation of long-term knowledge over time and mitigate catastrophic forgetting. Recently, some works have explored the effectiveness of using vision transformers (ViT) in incremental learning to improve the generality of learned knowledge, such as~\cite{wang2022continual,douillard2022dytox,mohamed2023d3former}. Particularly, Douillard et al.~\cite{douillard2022dytox} introduced an encoder-decoder architecture that dynamically expands task tokens to learn task-specific tokens and classifiers, which has proven successful in incremental learning. However, these ViT-based methods focus on enhancing the generality of enoder or decoder, not both, and therefore can limit  the adaptability of learned knowledge to new categories. Meanwhile, these methods do not handle class imbalance issues well.\par
To address this challenge, we integrate feature boosting~\cite{wang2022foster} into DyTox's  encoder-decoder architecture~\cite{douillard2022dytox} to develop  a two-stage training framework, dubbed SEDEG. In stage 1, SEDEG utilizes feature boosting to train an ensembled encoder, which obtains more generalized representations. By doing so, we are able to enhance the generality of the decoder and achieve a more balanced classifier when using these representations to train the decoder. In stage 2, SEDEG trains the new encoder using a balanced knowledge distillation (KD) strategy and feature KD to transfer knowledge from the ensemble encoder to the new encoder. Our experiments show that SEDEG effectively enhances the generality of both the decoder and the encoder, leading to significant improvements in performance.

In summary, this paper presents the following highlights:
 \begin{itemize}[leftmargin=*, topsep=0pt, itemsep=0pt, parsep=0pt]
     \item We adapt feature boosting by employing specially designed loss functions and integrate it into the encoder-decoder architecture of DyTox to create SEDEG.
     \item SEDEG is a novel ViT-based method that sequentially enhances the generality of the decoder and the encoder.
     \item Extensive experiments are conducted to showcase the exceptional performance of our approach on various continual learning benchmarks, and ablation studies validate the effectiveness of the components of SEDEG.
 \end{itemize}

 \section{Methodology}
 \subsection{Overview of SEDEG}
\begin{figure*}[t!]
    \centering
    \begin{subfigure}[t]{0.95\textwidth}
        \centering
        \includegraphics[width=0.7\linewidth]{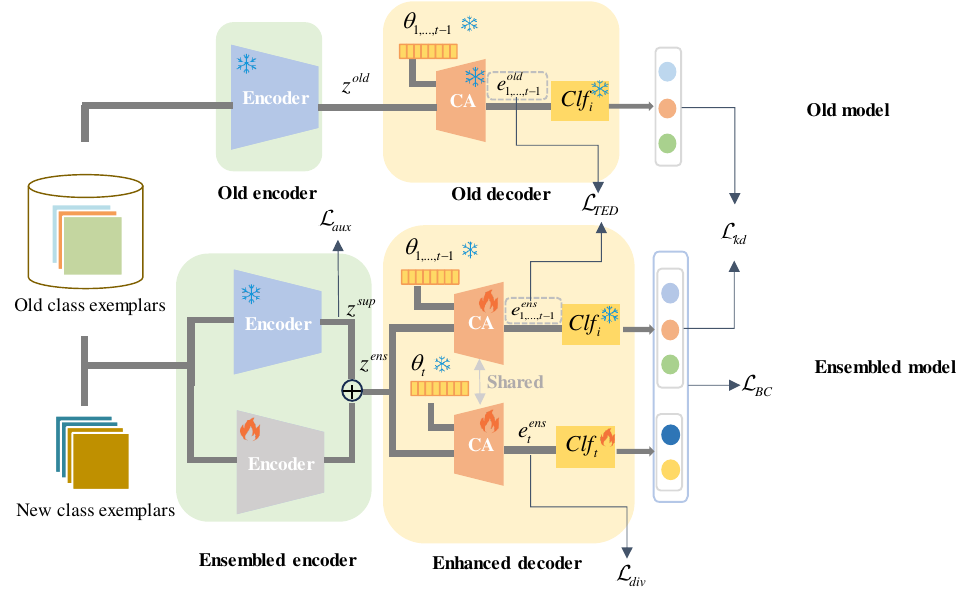}
        \caption{\textbf{Stage 1 of SEDEG.} SEDEG learns an enhanced and generalized decoder at this stage by configuring an ensembled encoder and training the ensembled encoder and the decoder jointly.}
        \label{subfig:phase1}
    \end{subfigure}
    \vspace{0.5cm}
    \begin{subfigure}[t]{0.95\textwidth}
        \centering
        \includegraphics[width=0.7\linewidth]{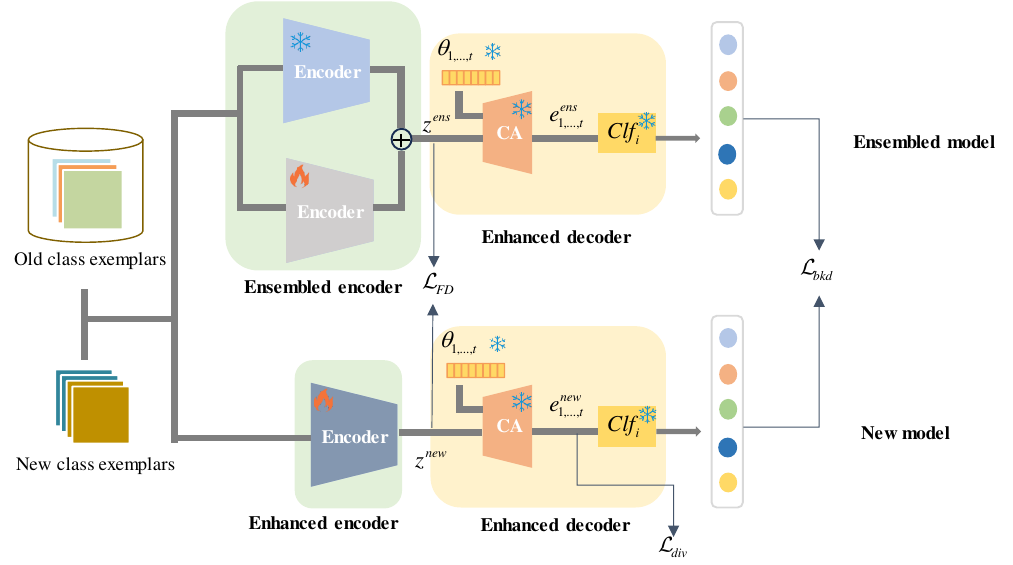}
        \caption{\textbf{Stage 2 of SEDEG.} SEDEG employs an effective knowledge distillation strategy to transfer the knowledge from the ensembled encoder to a new enhanced encoder.}
        \label{subfig:phase2}
    \end{subfigure}
    \caption{\textbf{Architecture of SEDEG.} Our SEDEG framework is composed of two stages. (a) Stage 1 learns an ensembled encoder and an enhanced decoder, and (b) stage 2 compresses the ensembled encoder into an enhanced encoder. Each of the two stages uses old class exemplars from the memory buffer as well as new class exemplars from the current task (task t). Following DyTox~\cite{douillard2022dytox}, all encoders represent a 5-layer self-attention block, CA represents a one-layer cross-attention block (TAB, Task Attention Block), ${\theta}_i$ represents the task token of task $i$ ($i \in \{1,...,t\}$), and ${Clf}_i$ stands for the classification head of task $i$.}
    \label{fig:overall-arc}
\end{figure*}

Our SEDEG framework is composed of two stages, as shown as Fig~\ref{fig:overall-arc}. The first stage (Fig ~\ref{subfig:phase1}) learns an ensembled encoder and an enhanced decoder, and the second stage compresses the ensembled encoder into an enhanced encoder. Specifically, the first stage introduces an additional encoder to supplement the features that the old encoder hasn't learned. The two encoders form an ensembled encoder to boost the features and helps to learn a generalization enhanced decoder. The second stage (Fig~\ref{subfig:phase2}) employs effective knowledge distillation methods to restore the parameter count of the old encoder while retaining the feature extraction capability of the ensembled encoder and learn an generalization enhanced encoder.

\subsection{Encoder Ensembled  and Decoder Enhancement}
To learn an enhanced and generalized decoder, we configure an ensembled encoder and trained the ensembled encoder and the decoder jointly. 
\subsubsection{Encoder Ensembled} We duplicate the encoder of the old model and freeze its parameters, adding a trainable encoder to supplement the features that the old encoder has not learned. The features learned by both encoders are fused by a simple and homogeneous channel-wise addition manner, and the fused features are then input into the decoder for training. To assist the supplementary encoder in learning the residual and boosted features for generalization across new and old tasks, we append an auxiliary classification head behind the supplementary encoder. This involves flattening the features outputted and adding a linear classification head. The classification involves all classes encountered thus far (new and old tasks) and the loss function is a binary cross-entropy loss following DyTox~\cite{douillard2022dytox}. This is presented as Equation (\ref{eq:aux-loss}), with $z^{sup}$ the output features of the supplementary encoder, $vec( \cdot )$ the flatten operation, $\phi(\cdot)$ the linear classification head, $o^{sup}$ the corresponding output logits and $y$ the ground truth.

\begin{equation}
\begin{split}
& o^{sup} = \phi(\text{vec}({z^{sup}})) \in \mathbb{R}^{\lvert \mathcal{Y}_{1:t}\rvert}, \\
& \mathcal{L}_{aux} = \text{BinaryCrossEntropy}(o^{sup}, y). \\
\end{split}
\label{eq:aux-loss}
\end{equation}

\subsubsection{Decoder Enhancement} In the decoder part, we maintain the parameters of task tokens and corresponding classification heads for old tasks frozen, while keeping the task token, the classification head of new task as well as  the cross-attention block trainable. This is essentially similar to DyTox, when a new task arrives. However, unlike DyTox, which independently uses an additional "fine-tuning" stage to downsample new task data to generate a balanced dataset and then fine-tune the entire model, we directly employ a balanced softmax classification~\cite{ren2020balanced} to train the model. The balanced classification $\mathcal{L}_{BC}$ is formulated as Equation (\ref{eq:bc-loss}), with with $o^{ens}_{j}$ ensembled model's class$-j$ output, $s_j$ number of samples in class $j$ and $\tau$ the temperature parameter set to 1. 

\begin{equation} 
\begin{split}
          & \mathcal{L}_{BC} = - \log \left( \frac{\exp(o^{ens}_{y} + \tau \log s_{y})}{\sum_{j=1}^{C} \exp(o^{ens}_{j} + \tau \log s_j)} \right).\\
\end{split}
\label{eq:bc-loss}
\end{equation}

To enhance the decoder's generalization capability on previous tasks and minimize forgetting of these tasks, we introduced \textbf{Task Embedding Distillation (TED)} and so prevent the decoder from forgetting task-level representations learned in previous tasks. TED compute a distance penalty between the task embeddings derived from the old model's TAB outputs and the corresponding outputs from the ensembled model. The TED loss function is shown as Equation (\ref{eq:ted-loss}), with $t$ the task number of current task (starting from 1 and $t>1$ ), $e_{i}^{\text{old}}$ task embedding on task $i$ of old model and $e_{i}^{\text{ens}}$ task embedding on task $i$ of ensembled model.

\begin{equation}
\mathcal{L}_{TED} = \frac{1}{t-1} \sum_{i=1}^{t-1} (e_{i}^{\text{old}} - e_{i}^{\text{ens}})^2.
\label{eq:ted-loss}
\end{equation}

\subsubsection{Losses} The loss function of the Encoder Ensembled  and Decoder Enhancement stage ($\mathcal{L}_1$) consists of 5 components, with two of them (Divergence loss
and Logits KD loss) identical to DyTox~\cite{douillard2022dytox}. To address the issue of class imbalance, we enhanced DyTox's binary cross-entropy classification function to a balanced classification loss function; To prevent decoder forgetting the task-level feature representations learned on historical tasks, we introduced the TED (Task Embedding Distillation) loss; Additionally, to assist the supplementary encoder in learning feature enhancements, we added an auxiliary classification loss afterward.
The total loss function is as Equation (\ref{eq:stage1-loss}),  with $\alpha$, $\lambda$ the same meaning and setting as DyTox, and $\mu$, $\xi$ hyperparameters set to 1.0, 0.1 correspondingly.
\begin{small}
\begin{equation} 
\begin{split}
              & \mathcal{L}_1 = (1 - \alpha)\mathcal{L}_{BC} + \alpha\mathcal{L}_{kd} + \lambda\mathcal{L}_{div} + \mu\mathcal{L}_{aux} + \xi\mathcal{L}_{TED}. \\
\end{split}
\label{eq:stage1-loss}
\end{equation}
\end{small}

\subsection{Encoder Enhancement} 
We utilize model compression techniques to restore the parameter count of the ensembled encoder to that of the old encoder, while preserving the feature extraction capabilities of the ensembled encoder, aiming to learn a generalized and enhanced new encoder. This involves employing knowledge distillation strategies to transfer the knowledge from the ensembled encoder to a new enhanced encoder. We freeze the decoder part (including all task tokens, all classification heads and task attention block) and use a Feature Distillation (FD) and a Balanced Logits Distillation (BLD) to achieve this. The FD aligns output features of ensembled encoder with the output features of enhanced encoder, thereby maintaining the feature extract capability of ensembled encoder. This is to enable the new enhanced encoder to keep the feature extraction capabilities of the ensembled encoder. The loss function is shown as Eqution (\ref{eq:fd-loss}), with $z^{ens}$ ensembled encoder output features, $z^{new}$ enhanced encoder output features and $\left\| \cdot \right\|_F$ denotes the Frobenius norm following PODNet~\cite{douillard2020podnet}.

\begin{equation}
\mathcal{L}_{FD} = \left\| z^{new} - z^{ens} \right\|_F.
\label{eq:fd-loss}
\end{equation}

Due to the severe class imbalance issue in the dataset used for distillation, it becomes challenging for the new model to learn the feature extraction capability on old tasks. We use a balanced KD~\cite{zhang2023balanced} to address this problem. The loss function is as Equation (\ref{eq:bld-loss}), with $\tau$ the temperature parameter set to 1, $\sigma(\cdot)$ the sigmoid function, $o^{new}_j$ the output logits of new model on class-$j$, $o^{ens}_j$ the output logits of ensembled model on class-$j$ and $w_j$ the per-class weight for the $j$-th class following FOSTER~\cite{wang2022foster}.

\begin{small}
\begin{equation}
\mathcal{L}_{BLD} = - \sum_{j=1}^{C} w_j \cdot \left( \sigma\left(\frac{o^{new}_{j}}{\tau}\right) \cdot \log\left(\sigma\left(\frac{o^{ens}_{j}}{\tau}\right)\right) \right).
\label{eq:bld-loss}
\end{equation}
\end{small}

\subsubsection{Losses} The loss function of the Encoder Enhancement stage ($\mathcal{L}_2$) consists of 3 components, with one of them (Divergence loss) identical to DyTox. The loss function of Encoder Enhancement stage is as Equation (\ref{eq:stage2-loss}), with $\beta$ hyperparameter set to 1.
\begin{equation} 
\begin{split}
              & \mathcal{L}_2 = \mathcal{L}_{BLD} + \lambda\mathcal{L}_{div} + \beta\mathcal{L}_{FE}. \\
\end{split}
\label{eq:stage2-loss}
\end{equation}

\section{Experiments and Results}
\subsection{Benchmarks \& implementation}
\noindent
\textbf{Benchmarks \& Metrics} We evaluated our proposed method on three scale-different classic continual learning datasets: CIFAR100~\cite{krizhevsky2009learning}, Tiny-ImageNet200~\cite{le2015tiny}, and ImageNet100~\cite{russakovsky2015imagenet}. We split CIFAR100 into 5 phases (20 new classes per phase), 10 tasks (10 new classes per phase) and 20 tasks (5 new classes per phase); split Tiny-ImageNet200 into 10 phases (20 new classes per phase) and split ImageNet100 into 10 tasks (10 new classes per phase). We report the "LAST" accuracy (the final accuracy after the last phase).

\subsection{Implementation}
To validate our approach across datasets with different input scales and reduce computational cost, we evaluate SEDEG on three datasets of different scale: CIFAR100, Tiny-ImageNet200 and ImageNet100. Hyperparameters related on input scale is shown as Table~\ref{tab:hyperparameters}. The main difference reflected in the model architecture across various datasets is primarily in the setting of the patch size. We conducted comparative experiments with various SOTA (State-Of-The-Art) methods in the same environment, with all training and inference performed on dual Tesla V100 GPUs and all reported results are averaged over three different natural class order. In order to make a fairer comparison with DyTox, we start training SEEDG on the second task and use the DyTox model trained on the first task as the old model.

\begin{table}
\caption{\textbf{SEDEG's hyperparameters for three datasets}. The main difference lies in the patch size on different input-scale datasets.}
\label{tab:hyperparameters}
\centering
\scalebox{0.8}{
\begin{tabular}{|c|c|c|c|}
\hline
\textbf{Hyperparameter} & \textbf{CIFAR-100} & \textbf{Tiny-ImageNet200} & \textbf{ImageNet100} \\
\hline
\# SAB &  & 5 &  \\
\# TAB &  & 1 &  \\
\# Attention Heads &  & 12 &  \\
Embed Dim &  & 384 & 3 \\
Input Size & 32 & \textbf{64} & 224 \\
Patch Size & 4 & \textbf{8} & 16 \\
\hline
\end{tabular}
}
\end{table}

\subsection{Results}
\noindent

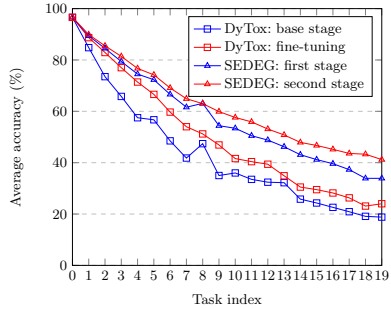
\begin{figure}[htbp]
\centering
\begin{tikzpicture}[scale=0.6]
\begin{axis}[
    xlabel={Task index},
    ylabel={Average accuracy (\%)},
    xmin=0, xmax=19,
    ymin=0, ymax=100,
    xtick={0,1,2,3,4,5,6,7,8,9,10,11,12,13,14,15,16,17,18,19},
    ytick={0,20,40,60,80,100},
    legend pos=north east,
    legend style={
        cells={anchor=west},
        font=\footnotesize
    },
    ymajorgrids=true,
    grid style=dashed,
]

\addplot[
    color=blue,
    mark=square,
    ]
    coordinates {
    (0,96.6)
    (1,84.8)
    (2,73.5)
    (3,65.8)
    (4,57.5)
    (5,56.7)
    (6,48.5)
    (7,41.8)
    (8,47.4)
    (9,35.0)
    (10,36.0)
    (11,33.5)
    (12,32.4)
    (13,32.2)
    (14,25.8)
    (15,24.3)
    (16,22.6)
    (17,20.9)
    (18,19.1)
    (19,18.8)
    };
   \addlegendentry{DyTox: base stage}

\addplot[
    color=red,
    mark=square,
    ]
    coordinates {
    (0,96.6)
    (1,88.7)
    (2,82.9)
    (3,77.0)
    (4,71.4)
    (5,66.6)
    (6,59.7)
    (7,54.0)
    (8,51.2)
    (9,46.9)
    (10,41.6)
    (11,40.4)
    (12,39.4)
    (13,34.9)
    (14,30.5)
    (15,29.5)
    (16,28.2)
    (17,26.3)
    (18,23.1)
    (19,24.0)
    };
    \addlegendentry{DyTox: fine-tuning}

\addplot[
    color=blue,
    mark=triangle,
    ]
    coordinates {
    (0,96.6)
    (1,89.4)
    (2,84.5)
    (3,79.4)
    (4,74.5)
    (5,72.3)
    (6,66.6)
    (7,61.6)
    (8,63.1)
    (9,54.4)
    (10,53.4)
    (11,50.5)
    (12,48.8)
    (13,46.2)
    (14,43.1)
    (15,41.2)
    (16,39.6)
    (17,37.3)
    (18,33.9)
    (19,33.9)
    };
   \addlegendentry{SEDEG: first stage}

\addplot[
    color=red,
    mark=triangle,
    ]
    coordinates {
    (0,96.6)
    (1,89.9)
    (2,85.5)
    (3,81.4)
    (4,76.7)
    (5,74.2)
    (6,69.1)
    (7,64.9)
    (8,63.0)
    (9,59.9)
    (10,57.6)
    (11,55.9)
    (12,53.1)
    (13,50.8)
    (14,47.9)
    (15,46.7)
    (16,45.2)
    (17,43.6)
    (18,43.3)
    (19,41.2)
    };
   \addlegendentry{SEDEG: second stage}
    
\end{axis}
\end{tikzpicture}
    \caption{Average accuracy at different stages of DyTox\cite{douillard2022dytox} and SEDEG with memory size 200 on CIFAR100. Our method (SEDEG) outperforms DyTox~\cite{douillard2022dytox} by a larger margin on both stages.}
    \label{fig:stage-level}
\end{figure}

\noindent
\textbf{Comparison with SOTA methods} We compare our method with SOTA methods in Class-Incremental Learning (CIL) scenario. 
The results are shown in Table \ref{tab:sota-results}. With last accuracy as the indicator, our method overall outperforms other methods by a large margin. For example, on the CIFAR100 dataset, with 5 tasks, the last accuracy is 58.09, a gain of 10.42 points compared to DyTox (the second best). With 10 tasks, the last accuracy is 53.71, exceeding DyTox by 21.09 points and outperform AMD (the second best) by 6.79 points. With 20 tasks, the last accuracy is 41.20, exceeding DyTox by 9.40 points and outperform SSIL (the second best) by 14.57 points. When the memory size is set to 500, SEEDG shows an average improvement of 5 points in last accuracy compared to DyTox, and outperform the second best result (D3Former) by 4 points overall. Similar substantial improvements are observed on the Tiny-ImageNet200 and ImageNet100 datasets as well. \\
\begin{table*}[h]
\caption{\textbf{Results on three datasets} averaged over three different natural class order. The best results are highlighted in \textbf{bold}, and the second best results are \underline{underlined}.}
\label{tab:sota-results}
\resizebox{\linewidth}{!}{
\begin{tabular}{|c|c|c|ccc|c|c|}
\hline
\multirow{2}{*}{\textbf{Buffer Size}} & \multirow{2}{*}{\textbf{Backbone}} & \textbf{Dataset} & \multicolumn{3}{c|}{\textbf{CIFAR-100}} & \textbf{Tiny-ImageNet} & \textbf{ImageNet-100} \\ \cline{3-8} 
                                      &                                     & \textbf{Phases}  & \textbf{5}       & \textbf{10}      & \textbf{20}      & \textbf{10}          & \textbf{10}          \\ \hline
\multirow{7}{*}{\textbf{200}}         & \multirow{5}{*}{CNN}               & DER++~\cite{yan2021der} & 28.36           & 21.43           & 14.76           & 12.43               & 12.32               \\ 
                                      &                                     & SSIL~\cite{ahn2021ss}   & 41.25           & 34.65           & \underline{26.63} & 32.78               & 25.92               \\ 
                                      &                                     & FOSTER~\cite{wang2022foster} & 35.62           & 29.87           & 22.44           & 10.84               & -                   \\ 
                                      &                                     & CKDF-iCaRL~\cite{li2022ckdf} & 44.83           & 35.70           & 22.58           & 24.63               & 34.38               \\ 
                                      &                                     & AMD~\cite{qiang2023mixture} & 45.67           & \underline{36.92} & 26.61           & 26.99               & 35.23               \\ \cline{2-8} 
                                      & \multirow{2}{*}{ViT}               & Dytox~\cite{douillard2022dytox} & \underline{47.67} & 32.62           & 23.96           & \underline{37.84}   & \underline{40.82}   \\ 
                                      &                                     & SEEDG (Ours)             & \textbf{58.09}   & \textbf{53.71}   & \textbf{41.20}   & \textbf{46.40}      & \textbf{52.54}      \\ \hline
\multirow{8}{*}{\textbf{500}}         & \multirow{5}{*}{CNN}               & DER++~\cite{yan2021der} & 39.13           & 36.09           & 20.97           & 19.64               & 14.61               \\ 
                                      &                                     & SSIL~\cite{ahn2021ss}   & 47.96           & 38.32           & 29.41           & 36.91               & 31.60               \\ 
                                      &                                     & FOSTER~\cite{wang2022foster} & 45.86           & 38.64           & 32.81           & 14.61               & 22.48               \\ 
                                      &                                     & CKDF-iCaRL~\cite{li2022ckdf} & 49.65           & 42.35           & 29.48           & 32.11               & 38.90               \\ 
                                      &                                     & AMD~\cite{qiang2023mixture} & 51.74           & 45.19           & 33.74           & 33.96               & 39.14               \\ \cline{2-8} 
                                      & \multirow{2}{*}{ViT}               & D3Former~\cite{mohamed2023d3former} & \underline{60.57} & \underline{52.19} & \underline{39.92} & 34.30               & 53.11               \\ 
                                      &                                     & Dytox~\cite{douillard2022dytox} & 58.93           & 48.54           & 37.16           & \underline{47.12}   & \underline{54.42}   \\ 
                                      &                                     & SEEDG (Ours)             & \textbf{64.29}   & \textbf{57.73}   & \textbf{44.68}   & \textbf{52.71}      & \textbf{56.88}      \\ \hline
\end{tabular}
}
\end{table*}
\textbf{Visualization} We recorded the performance of the model on the test dataset after the completion of different stages and results are shown as Fig~\ref{fig:stage-level}. Using average accuracy~\cite{rebuffi2017icarl} as indicator, SEDEG consistently outperformed DyTox by a significant margin at various stages.

\subsection{Ablation and analysis}
\subsubsection{Componets Abalation}
Because our method is divided into two stages and we have employed effective tricks in each stage to enhance the performance. To validate the effectiveness of the various techniques we proposed, we conducted ablation experiments on each stage separately. When conducting ablation experiments on one stage, the other stage should remain unchanged.\par
\textbf{Tricks on the first stage} During this stage, we applied the following tricks: 1) \textbf{Auxiliary loss} To help the new encoder learn both the task-specific features of new current task and the residual features of old tasks, we added an auxiliary loss ($\mathcal{L}_{aux}$) to it. The auxiliary loss is applied to all seen classes, including both the classes from old tasks and the classes from the new task, for performing full classification. 2) \textbf{Embeddings KD} To learn feature embeddings that are more representative for the new task, we introduced a distillation loss to the feature embeddings of the historical tasks. As a result, the learned feature embedding for the new task possesses global characteristics that distinguish from old tasks as well as adapt well to the new task. 3) \textbf{Balanced Classification Loss} To address the issue of class imbalance between the new task and old tasks (where the new task has a sufficient number of samples for each class while the rehearsal buffer contains a limited number of samples for each class of old tasks), we replace the original binary cross-entropy classification loss of Dytox with a balanced softmax classification loss.\\
We conducted experiments to validate the effectiveness of these tricks, and the experimental results are shown in Table \ref{tab4}. 

\textbf{Tricks on the second stage} During this stage, we applied the following tricks: 1) \textbf{Feature KD} Inspired by PODNet\cite{douillard2020podnet}, we incorporate feature distillation loss after the encoder to facilitate the feature learning capability distillation of the two-encoder teacher models. 2) \textbf{Balanced KD} The problem of class imbalance between new and old tasks can cause the student model to focus more on learning the feature extraction capability on the new task and neglect the feature learning on the old tasks. Therefore, we adopt balanced knowledge distillation loss to alleviate this issue. 3) \textbf{Distill Encoder Only} We only train the encoder of the student model while keeping its decoder parameters frozen, which has two benefits: On one hand, the student model can focus on learning the feature extraction capability of the teacher model without losing it due to adjustments in the decoder parameters, avoiding the risk of falling into local optima. On the other hand, this reduces the number of trainable parameters, making the training process easier and more likely to converge.\\
We conducted experiments to validate the effectiveness of these tricks, and the experimental results are shown in Table \ref{tab5}.

\begin{table}[H]
\centering
\begin{minipage}{0.48\textwidth}  
\caption{Ablation results for Auxiliary loss and Embeddings KD.}
\label{tab4}
\resizebox{\linewidth}{!}{
\begin{tabular}{|c|c|c|c|c|}
\hline
Auxiliary loss & Embeddings KD & Balanced Classification Loss & AVG & LAST \\
\hline
\checkmark & \checkmark & \checkmark & $68.80$ & $53.71$ \\
\checkmark & \checkmark &  & $64.21$ & $46.90$ \\
\checkmark &  & \checkmark & $68.25$ & $51.71$ \\
 &  & \checkmark & $67.76$ & $50.47$ \\
\hline
\multicolumn{3}{|c|}{DyTox} & $58.26$ & $32.62$ \\
\hline
\end{tabular}
}

\end{minipage}
\hfill  
\begin{minipage}{0.48\textwidth}
\caption{Ablation results for Feature KD and Balanced KD.}
\label{tab5}
\resizebox{\linewidth}{!}{
\begin{tabular}{|c|c|c|c|c|}
\hline
Feature KD & Balanced KD & Distill Encoder Only & AVG & LAST \\
\hline
\checkmark & \checkmark & \checkmark & $68.80$ & $53.71$ \\
\checkmark & \checkmark &  & $67.17$ & $53.04$ \\
\checkmark &  &  & $63.18$ & $40.62$ \\
 &  &  & $56.08$ & $30.10$ \\
\hline
\multicolumn{3}{|c|}{DyTox} & $58.26$ & $32.62$ \\
\hline
\end{tabular}
}

\end{minipage}
\end{table}

\subsubsection{Enhanced Encoder}
We utilized the t-SNE technique to reduce the dimensionality of high-dimensional data features output from new encoder of DyTox and SEDEG, visualizing on evaluation dataset of CIFAR-100 with 20 tasks. As shown in the Fig~\ref{fig:tsne-maps}, SEDEG effectively distinguishes clusters of different categories, including old tasks (category labels less than 5) and new tasks (category labels greater than 4), with each cluster being relatively concentrated and well-separated from others. In contrast, DyTox fails to effectively differentiate between categories, resulting in overlapping clusters.

\subsubsection{Enhanced Decoder}
 As there is no consistent representation across various tasks within the decoder, we don't directly visualize the performance of the decoder in terms of generalization. Rather, we conducted an ablation by freezing the entire encoder enhancement stage. Specifically, we compared our approach of training only the encoder (with the whole decoder frozen) against the method of fine-tuning the entire architecture. As shown in Table~\ref{tab5}, the performance achieved by freezing the entire decoder is superior to that of fine-tuning the whole architecture.

\section{Discussion and Analysis}
\subsection{\textbf{Class Imbalance vs. Memory Size}}
\noindent We also conducted experiments with a larger memory size, the results are shown as Table~\ref{tab6}. With a larger memory buffer size, the improvement effect of SEEDG to DyTox is relatively limited compared to when using a smaller memory, especially when there is a larger number of tasks. When the memory size is smaller, the stored historical samples are fewer, leading to a more pronounced class imbalance issue. Conversely, with a larger memory size, more historical samples can be stored, resulting in a less apparent class imbalance problem. Our method performs well in scenarios with low memory overhead, where only a small number of historical samples are stored, as we have effectively addressed the class imbalance issue.

\subsection{\textbf{Class Imbalance vs. Task Count}}
\noindent When the number of tasks is small, there are more categories within each task. With the same memory capacity, the number of samples per category that can be stored is relatively small, leading to a more pronounced class imbalance; vise versa. 
\begin{table}[h]
\caption{Results with large memory.} 
\label{tab6}
\centering

\resizebox{\linewidth}{!}{
\begin{tabular}{|c|c|llllll|ll|ll|}
\hline
\multirow{3}{*}{Buffer Size} & Dataset & \multicolumn{6}{c|}{CIFAR-100} & \multicolumn{2}{c|}{Tiny-ImageNet} & \multicolumn{2}{c|}{ImageNet-100} \\
                             & Phases  & \multicolumn{2}{c}{5} & \multicolumn{2}{c}{10} & \multicolumn{2}{c|}{20} & \multicolumn{2}{c|}{10} & \multicolumn{2}{c|}{10} \\
                             & ACC & AVG & LAST & AVG & LAST & AVG & LAST & AVG & LAST & AVG & LAST\\ \hline
\multirow{2}{*}{$1000$}       & Dytox & $73.01$ & $63.17$ & $71.37$ & $57.02$ & \textbf{68.51} & \textbf{50.94} & \textbf{51.75} & \textbf{40.57} & $67.68$ & $60.08$ \\
                             & SEEDG(Ours) & \textbf{74.85} & \textbf{66.53} & \textbf{73.04} & \textbf{58.11} & $66.72$ & $45.85$ & $51.66$ & $38.72$ & \textbf{68.57} & \textbf{60.94} \\ \hline
\multirow{2}{*}{$2000$}       & Dytox & $75.00$ & $66.67$ & $73.80$ & \textbf{61.27} & \textbf{72.25} & \textbf{56.15} & $52.96$ & \textbf{42.50} & $69.98$ & $63.56$ \\
                             & SEEDG(Ours) & \textbf{75.94} & \textbf{68.65} & \textbf{75.13} & $60.83$ & $70.90$ & $51.50$ & \textbf{53.44} & $41.42$ & \textbf{71.23} & \textbf{64.18} \\ \hline
\end{tabular}
}

\end{table}

\section{Conclusion}
\noindent In this paper, we proposed a novel two stage framework for continual learning termed Sequential Enhancement of Encoder and Decoder Generality (SEEDG). It begins by expanding the encoder to learn a more robust feature extractor, thereby adapting the decoder to enhance its generalization. Subsequently, we employ model compression techniques to restore the original parameter count, and train an enhanced and generalized encoder. The results show SEDEG can learn an adaptive and enhanced ViT for continual learning and achieve SOTA results in class incremental learning with limit memory cost. Future work will focus on enhancing the generalization capability of vision Transformers in scenarios where a large number of historical samples are cached, which enables our method to adapt to scenarios with varying memory overhead.


%
%
%
%
\bibliographystyle{splncs04}
\bibliography{reference}

\end{document}